\documentclass{bmvc2k}

\title{Large Scale Joint Semantic Re-Localisation and Scene Understanding via Globally Unique Instance Coordinate Regression}

\addauthor{Ignas Budvytis}{ib255@cam.ac.uk}{12}
\addauthor{Marvin Teichmann}{mttt2@cam.ac.uk}{12}
\addauthor{Tomas Vojir}{tv278@cam.ac.uk}{12}
\addauthor{Roberto Cipolla}{rc10001@cam.ac.uk}{1}

\addinstitution{
 Department of Engineering\\
 University of Cambridge\\
 Cambridge, UK
}
\addinstitution{
 indicates equal contribution
}

\runninghead{Budvytis, Teichmann, Vojir, Cipolla}{Semantic re-localisation}

\begin{document}

\maketitle
\vspace{-0.3cm}
\begin{abstract}
        In this work we present a novel approach to joint semantic localisation and scene understanding. Our work is motivated by the need for localisation algorithms which not only predict 6-DoF camera pose but also simultaneously recognise surrounding objects and estimate 3D geometry. Such capabilities are crucial for computer vision guided systems which interact with the environment: autonomous driving, augmented reality and robotics. In particular, we propose a two step procedure. During the first step we train a convolutional neural network to jointly predict per-pixel globally unique instance labels~\cite{BudvytisSC18} and corresponding local coordinates for each instance of a static object (e.g. a building). During the second step we obtain scene coordinates~\cite{shottonscenecoords13} by combining object center coordinates and local coordinates and use them to perform 6-DoF camera pose estimation. We evaluate our approach on real world (CamVid-360) and artificial (SceneCity) autonomous driving datasets~\cite{BudvytisSC18}. We obtain smaller mean distance and angular errors than state-of-the-art 6-DoF pose estimation algorithms based on direct pose regression~\cite{PoseNetKendallGC15,KendallC17} and pose estimation from scene coordinates~\cite{BrachmannR18} on all datasets.  Our contributions include: (i) a novel formulation of scene coordinate regression as two separate tasks of object instance recognition and local coordinate regression and a demonstration that our proposed solution allows to predict accurate 3D geometry of static objects and estimate 6-DoF pose of camera on (ii) maps larger by several orders of magnitude than previously attempted by scene coordinate regression methods~\cite{shottonscenecoords13,Brachmann2017DSACD,BrachmannR18,liangularscr18}, as well as on (iii) lightweight, approximate 3D maps built from 3D primitives such as building-aligned cuboids. \vspace{-0.2cm}
\end{abstract}

%-------------------------------------------------------------------------
\section{Introduction}
\begin{figure*}[ht]
\includegraphics[width=1.0\linewidth]{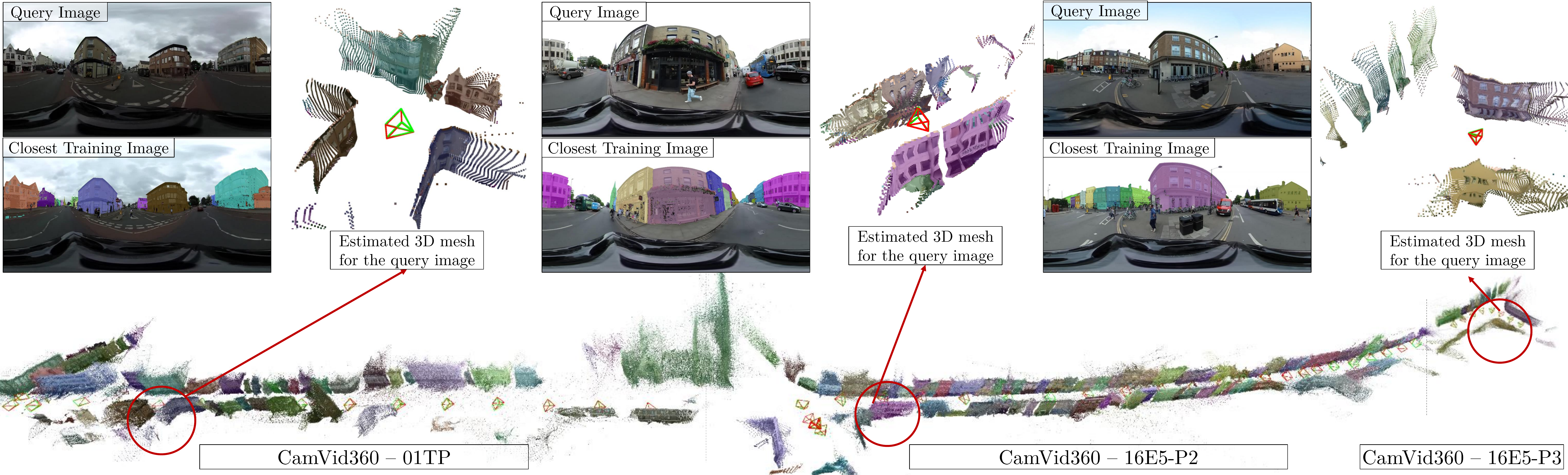}\vspace{-0.2cm}
   \caption{\small{Three triplets of images on the top of this figure illustrate a typical result of our framework for joint semantic re-localistation and scene understanding via globally unique instance coordinate prediction. The top left image of the triplet corresponds to a query image provided as an input to our network. The output of the network consists of per-pixel 3D coordinates and corresponding globally unique instance labels are shown on the right together with ground truth (green) and estimated (red) camera poses. The image at the bottom left shows the closest image in the database with ground truth building instance label images overlaid. The bottom part of the figure illustrates cumulative 3D point cloud built from predicted scene coordinates as well as ground truth (green) and estimated (red) camera poses for sequences 16E5-P2, 16E5-P3 and 01TP. See Section~\ref{secexperiments} for more quantitative and qualitative results. Zoom in for a better view.}}
   \vspace{-0.4cm}
\label{figintro}
\end{figure*}

As computer vision enabled robotic systems are increasingly deployed in the real world, simplicity, efficiency, verifiability and robustness of computer vision algorithms become highly important aspects of their design. An example of a desired solution satisfying the aforementioned requirements would likely include training a single network which would predict a structured, semantically meaningful output, the correctness of which could be verified at test time and from which all the necessary tasks for navigation and interaction with environment could be performed.

In this work we propose such a structured representation from which tasks of semantic segmentation, recognition and localisation can be performed efficiently and accurately. Our proposed solution is inspired by works on globally unique instance segmentation~\cite{BudvytisSC18} and scene coordinate regression~\cite{shottonscenecoords13,Brachmann2017DSACD,BrachmannR18}. It includes the following key steps. First, a dataset of densely sampled images, ideally a video, of the environment is created. It is labelled with globally unique instance labels~\cite{BudvytisSC18} and a corresponding 3D point cloud is obtained by running a structure-from-motion algorithm~\cite{mappilaryopensfm} on the collected images.  Second, a CNN is trained to simultaneously predict globally unique instance labels and local coordinates of corresponding objects. Finally, at test time scene coordinates are formed by combining object center coordinates with local coordinates for an 6-DoF camera pose estimation which is formulated as a solution to a perspective-n-point problem (PnP)~\cite{lepetitepnp,kneipupnp14}. See Figure~\ref{figintro} for an example output of our method.

We evaluate our approach on real world and artificial autonomous driving datasets. Our method predicts more than $53\%$ and $39\%$ of pixels within $50 cm$ of ground truth location for CamVid-360~\cite{BudvytisSC18} and SceneCity Medium~\cite{BudvytisSC18} datasets spanning approximately $1.5km$ and $11.5 km$ in driving length. We obtain 22 cm and 20 cm median distance error as well as $0.71^{\circ}$ and $0.76^{\circ}$ median angular errors on estimated camera poses for the same datasets. Our method outperforms competing deep learning based localisation methods based on either direct 6-DoF pose prediction~\cite{PoseNetKendallGC15,KendallC17} or pose estimation from scene coordinates~\cite{BrachmannR18} on all datasets. When tested on highly challenging scenarios of using a different camera (Google StreetView images) or re-localising in scenes with missing buildings~\cite{BudvytisSC18} our method demonstrates higher robustness than alternative approaches. Our contributions include: (i) a novel formulation of scene coordinate regression as two separate tasks of object instance recognition and local coordinate regression and a demonstration that our proposed solution allows to predict accurate 3D geometry of static objects and estimate 6-DoF pose of camera on (ii) maps larger by several orders of magnitude than previously attempted~\cite{shottonscenecoords13,Brachmann2017DSACD,BrachmannR18,liangularscr18}, as well as on (iii) lightweight, approximate 3D maps built from 3D primitives. 

The rest of this work is divided as follows. Section~\ref{secrelatedwork} discusses relevant work in localisation. Section~\ref{secmethod} provides details of our proposed localisation approach. Sections~\ref{secexperimentsetup} and~\ref{secexperiments} describe the experiment setup and corresponding results.
\vspace{-0.2cm}
\section{Related Work} \label{secrelatedwork} %\vspace{-0.2cm}

\begin{figure*}[t]
\begin{center}
   \includegraphics[width=1.0\linewidth]{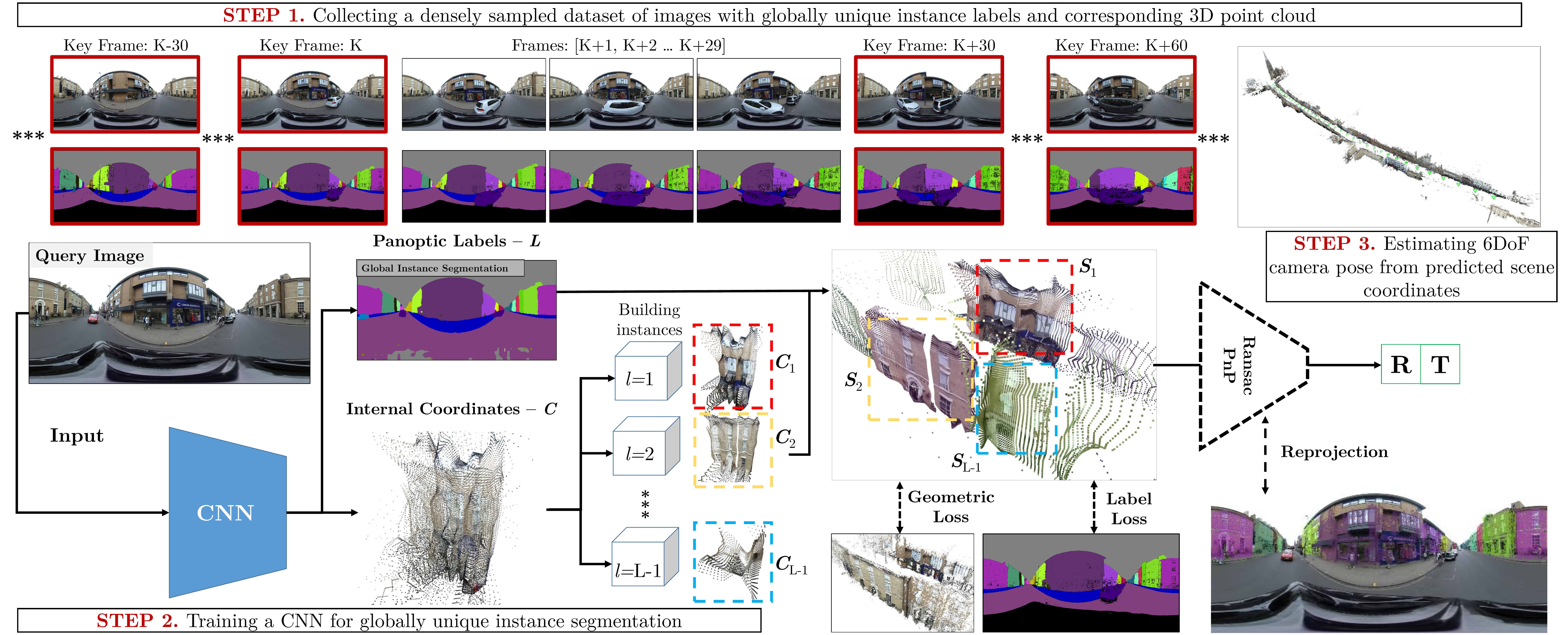} \vspace{-0.8cm}
\end{center}
   \caption{\small{This figure illustrates the three key steps of our method. First, a densely sampled dataset of images is collected and annotated with both class and globally unique instance labels. From these images a 3D point cloud is build using state-of-the-art library for structure from motion - OpenSFM~\cite{mappilaryopensfm}. Second, a CNN is trained to predict panoptic labels $L$ as well as local PCA whitened coordinates $C$ for each object instance. During the final step predicted local coordinates are unwhitened to obtain corresponding scene coordinates $S$. EPnP~\cite{lepetitepnp} with RANSAC is used for camera pose estimation. }} \vspace{-0.3cm}
\label{fignetwork}
\end{figure*}

In this section we provide a discussion of various related work on localisation.

\noindent \textbf{Matching based localisation.} Traditionally, large scale localisation problems are formulated as correspondence problems in the image domain or in the 3D point cloud domain. The first group of approaches work by identifying the most similar looking image in a database primarily in two ways: by employing either (i) a pipeline of keypoint detection and matching~\cite{Lowe2004,LIFTYi16,GoogleLandmarks} or (ii) fast-to-compare image level encoding~\cite{NetVladArandjelovic16,DenseVlad}. In order to obtain a 6-DoF pose estimation they are augmented with an additional step of establishing feature matches for one or more neighbour images and solving a perspective-n-point problem~\cite{kneipupnp14} inside a RANSAC~\cite{RANSACFischler} solver. The second group of approaches obtain local 3D geometry by using 3D sensors such as structured light~\cite{Izadi11kinectfusion,Scharstein2003}, time-of-flight~\cite{rwolcott2014a} cameras as well as RGB based structure from motion~\cite{torii} and match it to a pre-built 3D model of the environment. While both types of works have a potential for providing high accuracy pose estimates at large scale, they are limited by large storage requirements of feature indices or 3D point clouds and relatively slow correspondence estimation procedures.

\noindent \textbf{Direct location prediction.}
The need for test-time storage and correspondence estimation is addressed by the works which attempt to directly predict either a coarse~\cite{PlaNetWeyand2016} location or a full 6-DoF camera pose~\cite{PoseNetKendallGC15,KendallC17}. An estimation of location or a precise camera pose is obtained by simply training a deep network with a corresponding objective. The coarse methods~\cite{PlaNetWeyand2016} still require performing additional local feature matching if a 6-DoF pose estimate is needed and hence are not very efficient at test-time. In contrast, methods which directly predict camera pose demonstrate test-time efficiency as only a single pass through a network is required. However they are prone to over-fitting to the training images (e.g. a network may learn to predict a location based on the presence of a parked car in the image) and are not robust to changes in the environment as shown in Section~\ref{secexperiments} and discussed in detail in~\cite{torstenlimitations}.

\noindent \textbf{Localisation via scene coordinate prediction.} Test time robustness is increased by approaches which perform localisation via scene coordinate regression~\cite{shottonscenecoords13,Brachmann2017DSACD,BrachmannR18,liangularscr18}. Such works often train a per pixel 3D scene coordinate regressor whether using a CNN~\cite{Brachmann2017DSACD} or other method (e.g. Random Forest~\cite{shottonscenecoords13}) and solves a perspective-n-point problem to obtain the estimate of the camera pose. Early works focus on learning outlier masks~\cite{shottonscenecoords13} in order to remove unreliable candidates for pose estimation and propose a differential pose estimation~\cite{Brachmann2017DSACD} to be compatible with fully end-to-end training schemes at the expense of more complex learning task. In contrast, \cite{BrachmannR18} proposes to simplify the trainable components by making scene coordinate as the only trainable part of the algorithm. We further simplify their method by replacing the differentiable pose estimation algorithm with a classical one~\cite{kneipupnp14} and simply relying on our network ability to accurately predict 3D coordinate predictions. We also reformulate scene coordinate regression as a task of joint globally unique instance segmentation and prediction of local object coordinates (see Section~\ref{secmethod}) which allows us to obtain accurate pose estimates on orders of magnitude larger maps than in~\cite{shottonscenecoords13,Brachmann2017DSACD,BrachmannR18,liangularscr18} despite using training data consisting only of videos traversing environments of interest following a simple trajectory once. 

\noindent \textbf{Semantic localisation.} Semantic information is often incorporated into localisation frameworks in one of the two ways. Approaches of the first type perform keypoint filtering~\cite{OB17} or feature reweighting~\cite{kim2017crn,SemanticVisLoc} of dynamic or difficult objects. Approaches of the second type attempt an explicit fitting of 3D models of individual rooms~\cite{satkinbmvc2012}, or buildings~\cite{indooroutdooreccv16} or of detailed maps~\cite{SanFranLandmark,SanFranAlignment}.  The former methods often increase the accuracy of underlying localisation algorithms but do not directly address their robustness under changes in the environment. The latter methods are often slow at test time and are more suitable for data collection. A recent work of~\cite{BudvytisSC18} attempt to predict a rich representation of per-pixel globally unique instance labels and show that it is enough to perform localisation from it under severe changes in the environment. Our work augments this representation with local coordinate prediction which allows us to obtain 6-DoF pose estimates as opposed to performing image retrieval an to introduce robustness to unseen translation of the camera poses at test time as well as to avoid a computationally expensive step of explicit rotational alignment of label images.

\vspace{-0.3cm}
\section{Method}\label{secmethod} %\vspace{-0.3cm}

\begin{figure*}[t]
\begin{center}
\begin{tabular}{cc}
  &\hspace{-7.3cm}\footnotesize{(a)}
  \includegraphics[width=0.96\linewidth]{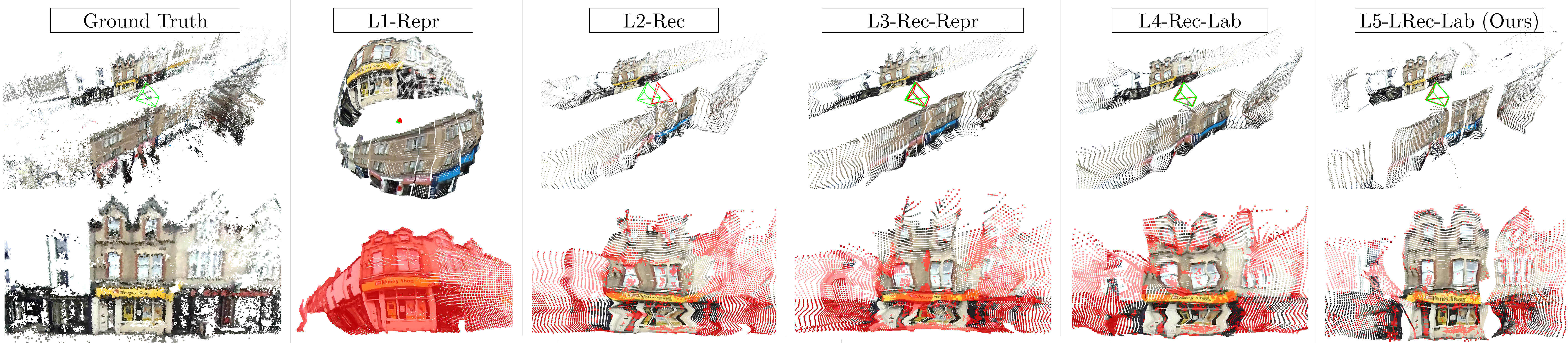} \vspace{-0.2cm} \\ 
 \hspace{-0.2cm}\footnotesize{(b)} 
 \hspace{-0.1cm}\includegraphics[width=0.5\linewidth]{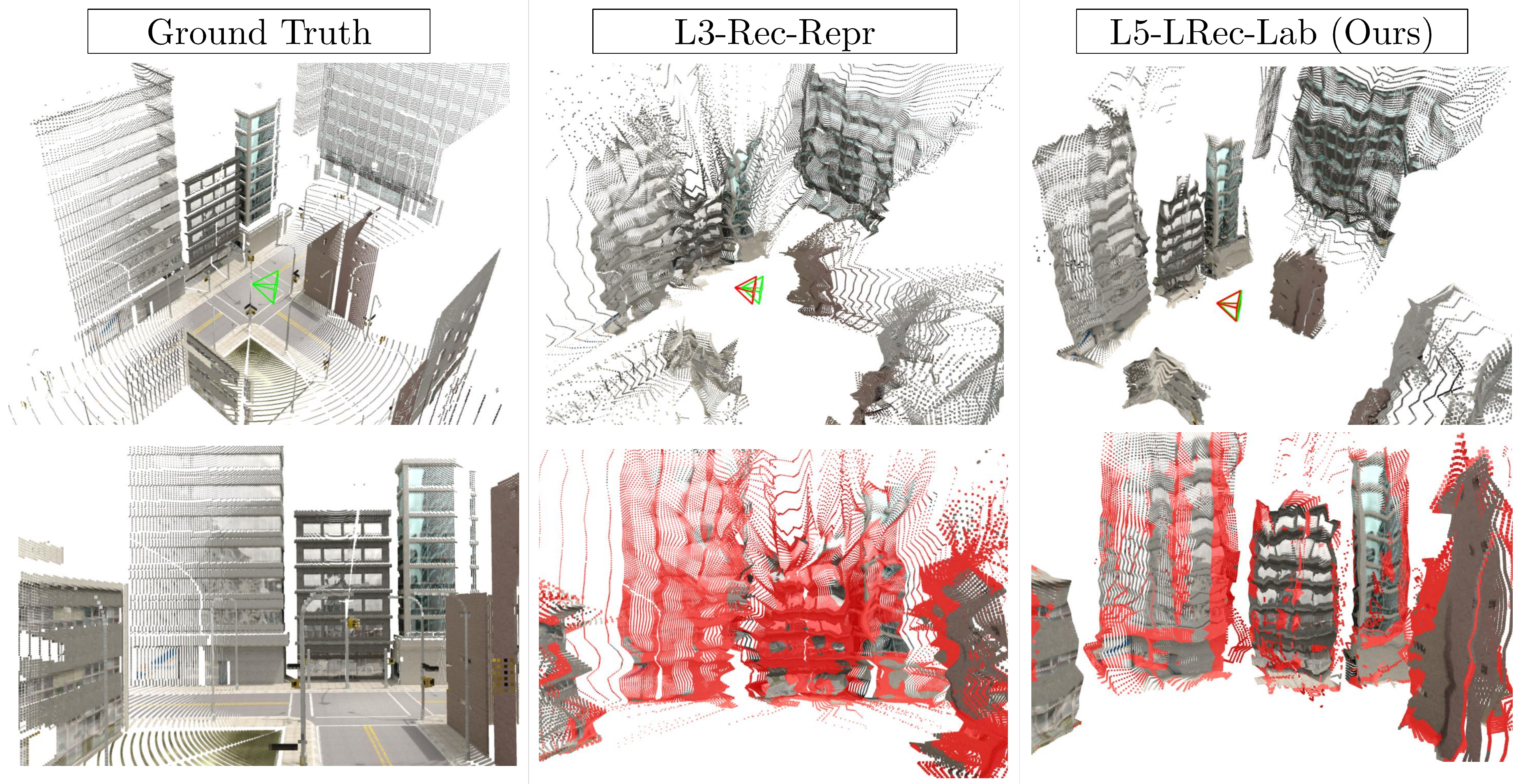} & \hspace{-0.3cm}\footnotesize{(c)}
 \includegraphics[width=0.43\linewidth]{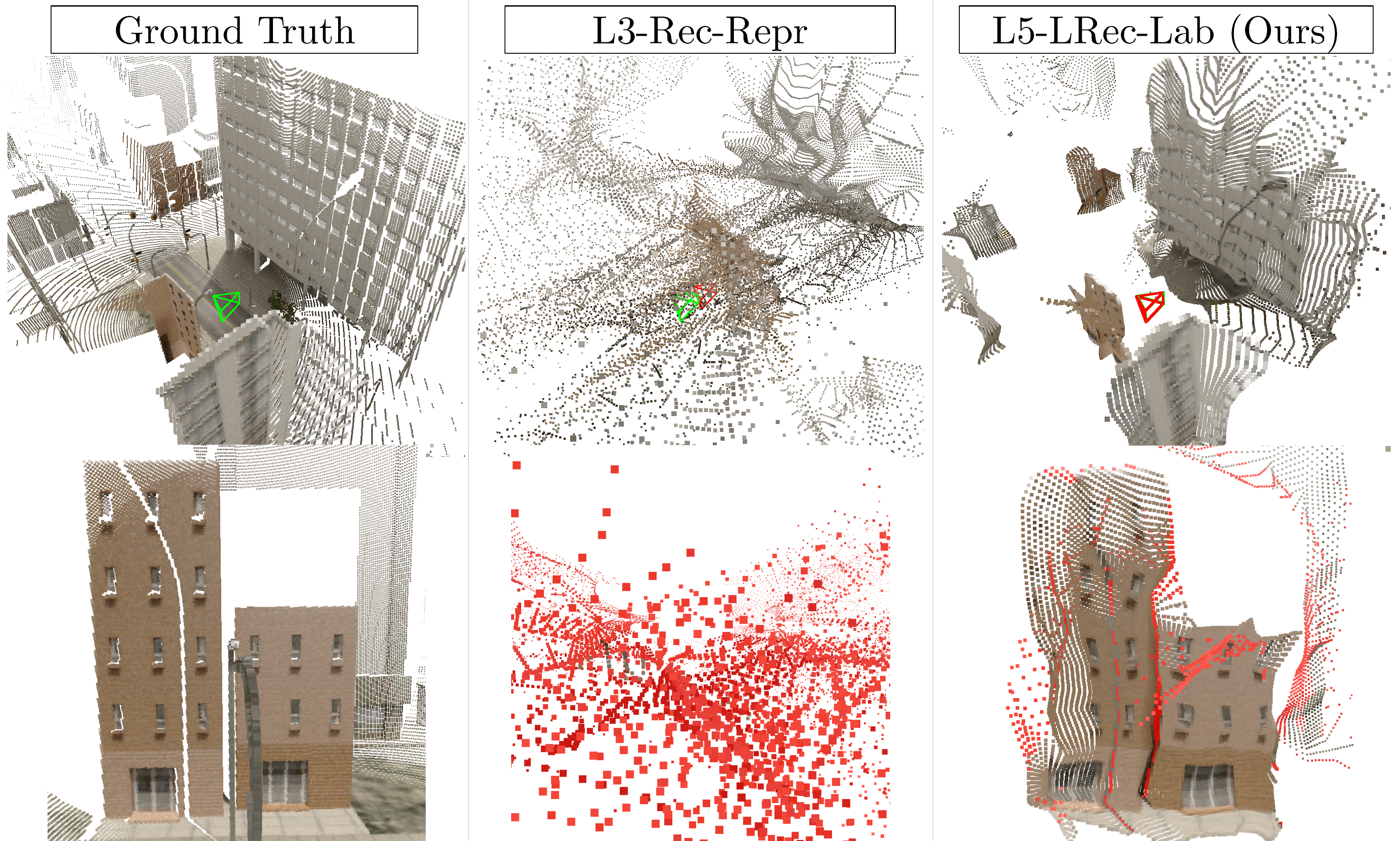}
\end{tabular}
\end{center} \vspace{-0.1cm}
   \caption{\small{Part (a) illustrates scene coordinates predicted by CNNs trained on CamVid-360 16E5-P2 sequence with one of five different losses (see Section~\ref{secexperiments} for definitions) for a single image. Two different views of the same point cloud are shown on top and bottom rows. Red pixels in the images on the bottom row indicate pixels for which predicted 3D location is further than 0.5m (a) and 1m (b,c) away from their true location. Parts (b) and (c) similarly illustrate scene coordinate predictions on Small and Large SceneCity datasets respectively. All networks except ones trained with $\textit{L1-Repr}$ loss learn a good approximation of scene geometry for CamVid-360 16E5-P2 sequence. However when map size increases (e.g. Large SceneCity dataset) the 3D reconstruction accuracy significantly decreases for methods which attempt to predict 3D values directly (e.g. \textit{$\text{L3-Rec-Repr}$}). Zoom in for a better view.}} \vspace{-0.4cm}
\label{fig3dqualitative}
\end{figure*}

Our proposed localisation framework consists of three key steps: data collection, training of a CNN to predict globally unique instance coordinates and pose estimation. More details are provided below and in Figure~\ref{fignetwork}.

\noindent \textbf{Data collection.} First, a densely sampled collection of panoramic images of the environment is obtained. Second, a subset of images (e.g. every 30 frames) are hand labelled with both class labels (e.g. sky, road, pedestrian) and globally unique instance labels of buildings\footnote{Note that instances of other static objects such as trees or road signs could also be used as demonstrated in~\cite{BudvytisSC18}.} and a label propagation algorithm~\cite{Budvytis2017ICCV} is used to label the rest of the images. Finally, camera pose estimates and corresponding semantic 3D point cloud are obtained using OpenSFM~\cite{mappilaryopensfm}, an open source structure from motion (SfM) library. Default parameter settings are used unless stated otherwise. The point cloud is projected for each training image using ground truth camera pose in order to produce a 3 channel image containing (x,y,z) coordinates of projected 3D points. When multiple 3D points are projected to the same pixel, the closest one with the same instance label as the source pixel is chosen.

\noindent \textbf{Training.} During the second step a CNN is trained to jointly predict (i) panoptic labels consisting of class labels (e.g. 10 class labels such as road, sky, people as in~\cite{BudvytisSC18}) and instance labels of buildings as well as (ii) local PCA whitened coordinates of building instances for each pixel. Unwhitening transformation for point $p$ with label $l$ is performed as follows: $S_{p}=W_{l}C_{p}+M_{l}$, where $C_{p}$ and $S_{p}$ denote correspondingly local whitened coordinates and scene coordinates of point $p$. $M_{l}$ is the mean coordinate of all points of label $l$ in the training data. $W_{l}$ is the whitening matrix for label $l$. We apply standard cross entropy loss for both class and instance labels as in~\cite{BudvytisSC18} and a euclidean distance loss for fitting whitened local instance coordinates $C_{p}$. Note that for each pixel a $3+L$ dimensional vector is predicted where 3 corresponds to a 3-dimensional coordinate and $L$ corresponds to a total number of panoptic labels. See Section~\ref{secexperimentsetup} for more details.

\noindent \textbf{Pose estimation.} Camera pose is estimated with EPnP~\cite{lepetitepnp} perspective-n-point solution with RANSAC~\cite{RANSACFischler} loop from predicted scene coordinates. Other standard solutions~\cite{kneipupnp14} to PnP can be used as well. RANSAC is run for 1000 iterations with points within $0.22^{\circ}$ considered as an outlier threshold. Since we aim to recover as accurate 3D coordinates as possible we do not consider employing differentiable pose estimation algorithms used in~\cite{Brachmann2017DSACD,BrachmannR18}. The reprojection loss component employed in such algorithms reduces the accuracy of predicted 3D geometry in favour of a more accurate camera pose estimation. This is demonstrated in Figure~\ref{figcamvidvariationsquantitative} where scene coordinate prediction accuracy is lower for the loss \textit{$\text{L3-Rec-Repr}$} ($40\%$ of points lie within $0.5m$ of ground truth) which combines reconstruction and reprojection losses than \textit{$\text{L2-Rec}$} ($44\%$ of points lie within $0.5m$ of ground truth) which directly aims at minimising euclidean distance between predicted and ground truth scene coordinates.  

\vspace{-0.3cm}
\section{Experiment Setup}\label{secexperimentsetup} %\vspace{-0.1cm}
Below we describe the details of the datasets, network architecture and evaluation protocol.

\noindent \textbf{CamVid-360 dataset.} CamVid-360~\cite{BudvytisSC18} is a dataset of panoramic videos captured by cycling along the original path of CamVid~\cite{Brostow2009}. CamVid-360 training set consists of 7835 images sampled at 30 fps, at resolution $1920\times960$ which cover sequences 016E5, 001TP of the original CamVid~\cite{Brostow2009} dataset. Query set contains both test sequence\footnote{ Note that unlike~\cite{BudvytisSC18} we do not use images from sequence 006R0 as this sequence is not covered in training data.} used in~\cite{BudvytisSC18} (318 images sampled at 1 fps) as well as a new additional test sequence obtained by downloading Google StreetView panoramic images along the tracks of the original dataset.  We estimate ground truth poses for testing images by minimising the reprojection errors of SIFT~\cite{Lowe2004} feature matches from 80 closest images in training dataset via robust EPnP~\cite{lepetitepnp}.

\noindent \textbf{SceneCity dataset.} SceneCity~\cite{BudvytisSC18} dataset contains images rendered from two artificial cities. See Figures~\ref{fig3dqualitative}(b,c) and~\ref{figreconstructedpaths} for example images and maps. The first city, referred as Small SceneCity, is borrowed from~\cite{ZhangRFS16}. It contains 102 buildings and 156 road segments. The second city, referred as Large SceneCity, contains 827 buildings and 966 road segments in total. Training database consists of 1146 and 6774 images sampled uniformly from each city respectively. 3D point clouds are obtained from Blender directly. Our algorithms are evaluated on two variants of Small SceneCity. For the first variant 300 camera poses are sampled uniformly from the original track of~\cite{ZhangRFS16}. For the second variant same camera poses are used but random $20\%$ of buildings are removed from the Small SceneCity map.  Query set for the Large SceneCity consists of 1000 samples near the center of road segments as explained in~\cite{BudvytisSC18}.

\begin{figure*}[t]
   \begin{tabular}{cccc}
  \hspace{-0.35cm}\footnotesize{(a)}\hspace{-0.03cm}\includegraphics[width=0.23\linewidth]{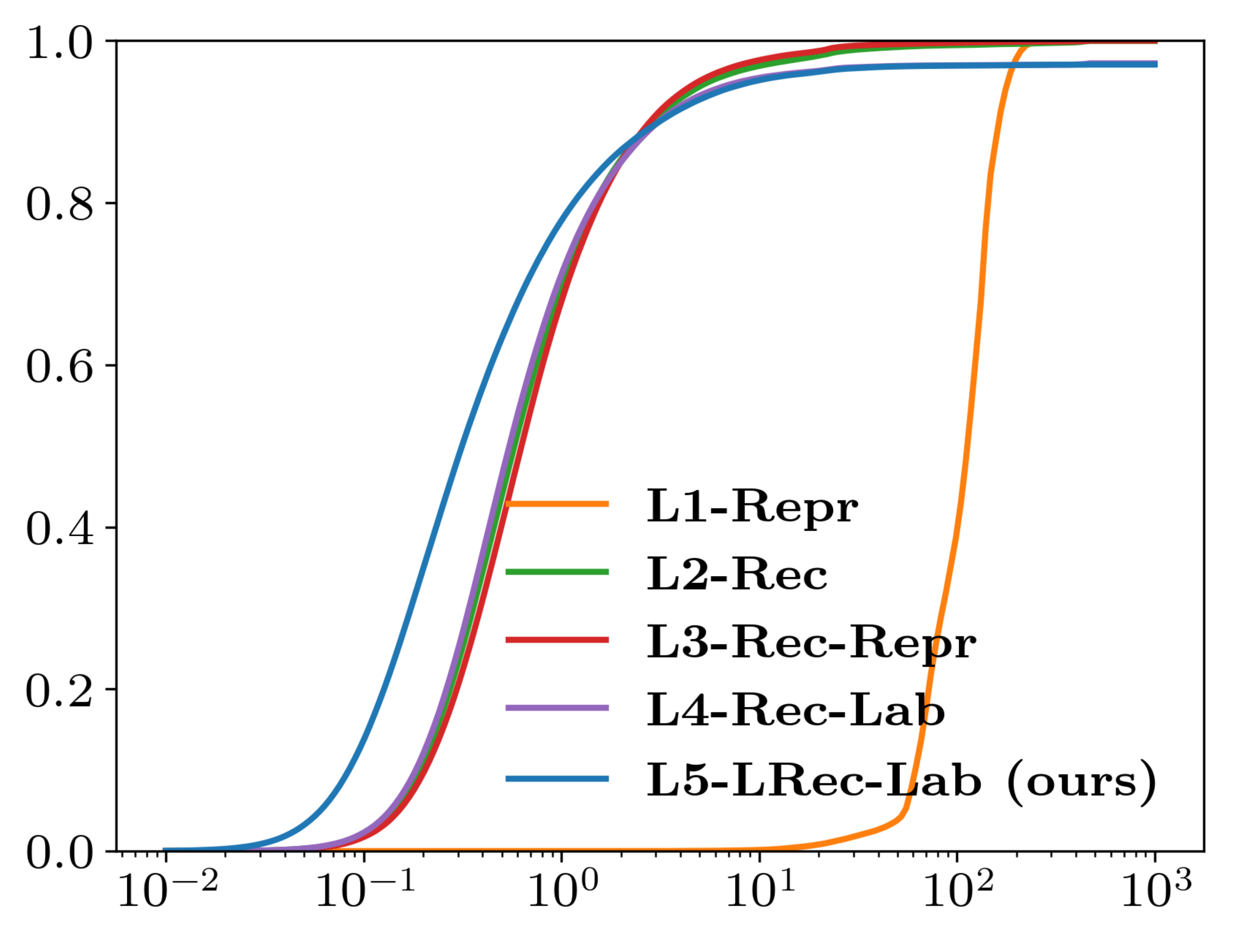}&
  \hspace{-0.35cm}\footnotesize{(b)}\hspace{-0.03cm}\includegraphics[width=0.23\linewidth]{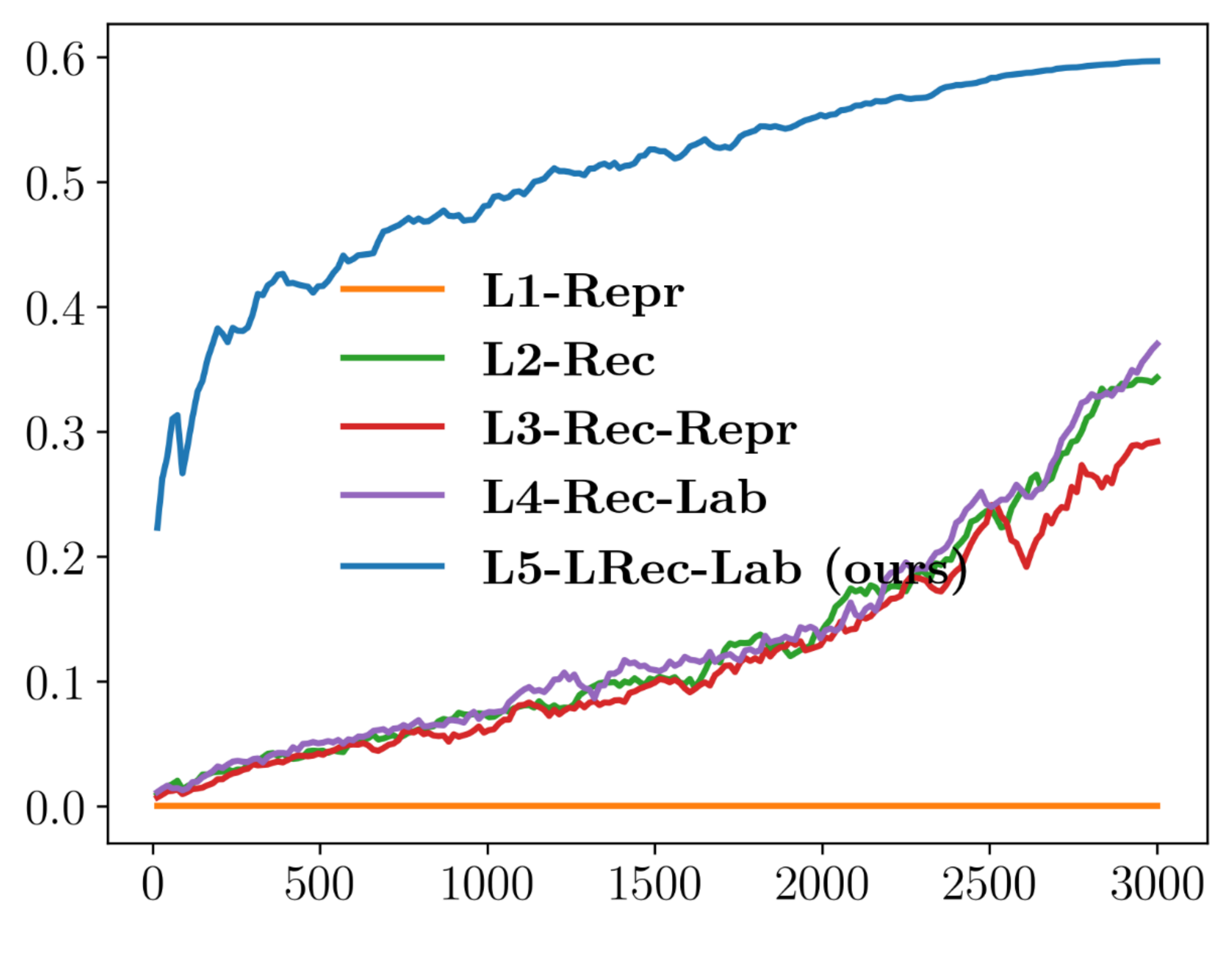}&
  \hspace{-0.35cm}\footnotesize{(c)}\hspace{-0.03cm}\includegraphics[width=0.23\linewidth]{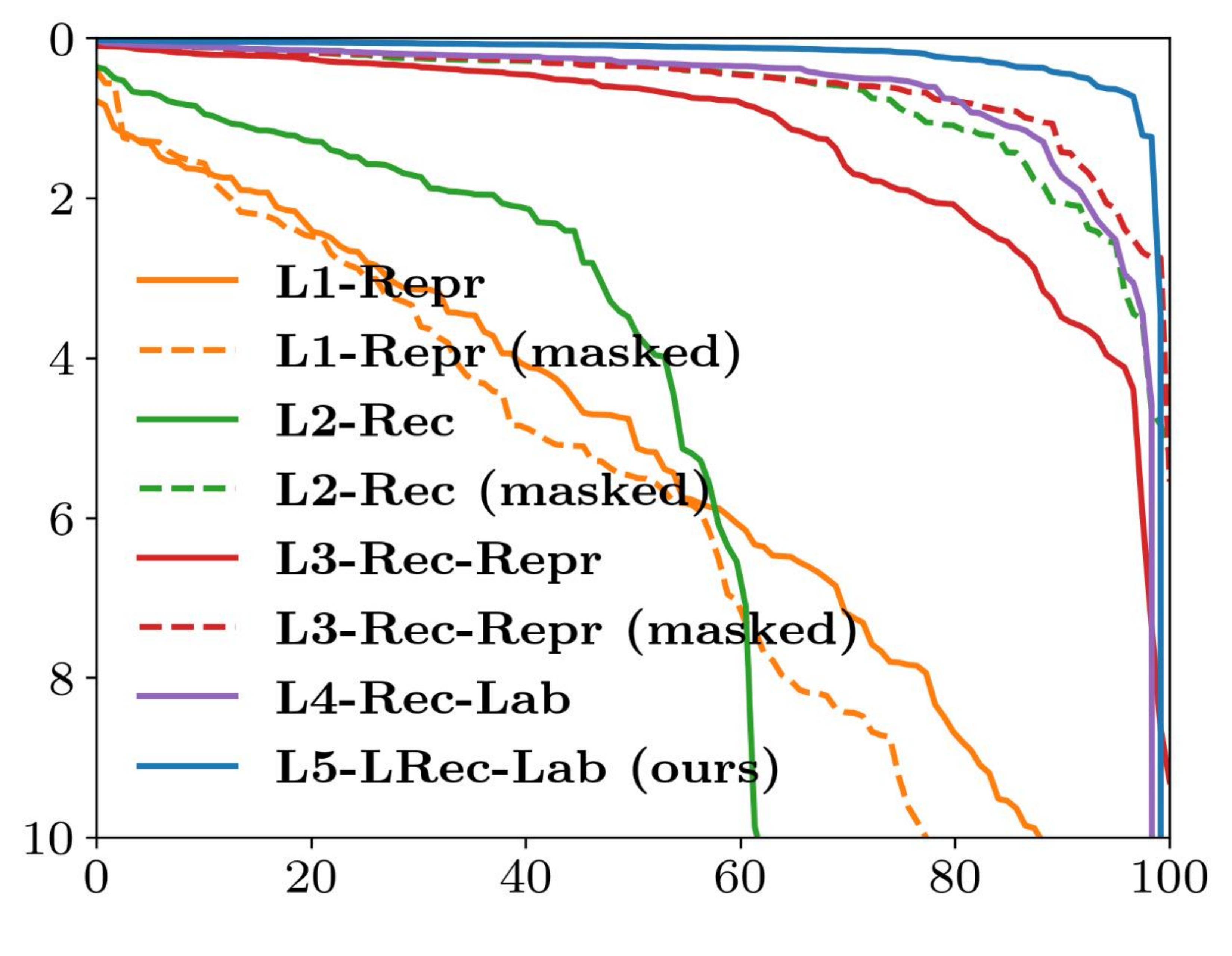}&
  \hspace{-0.35cm}\footnotesize{(d)}\hspace{-0.03cm}\includegraphics[width=0.23\linewidth]{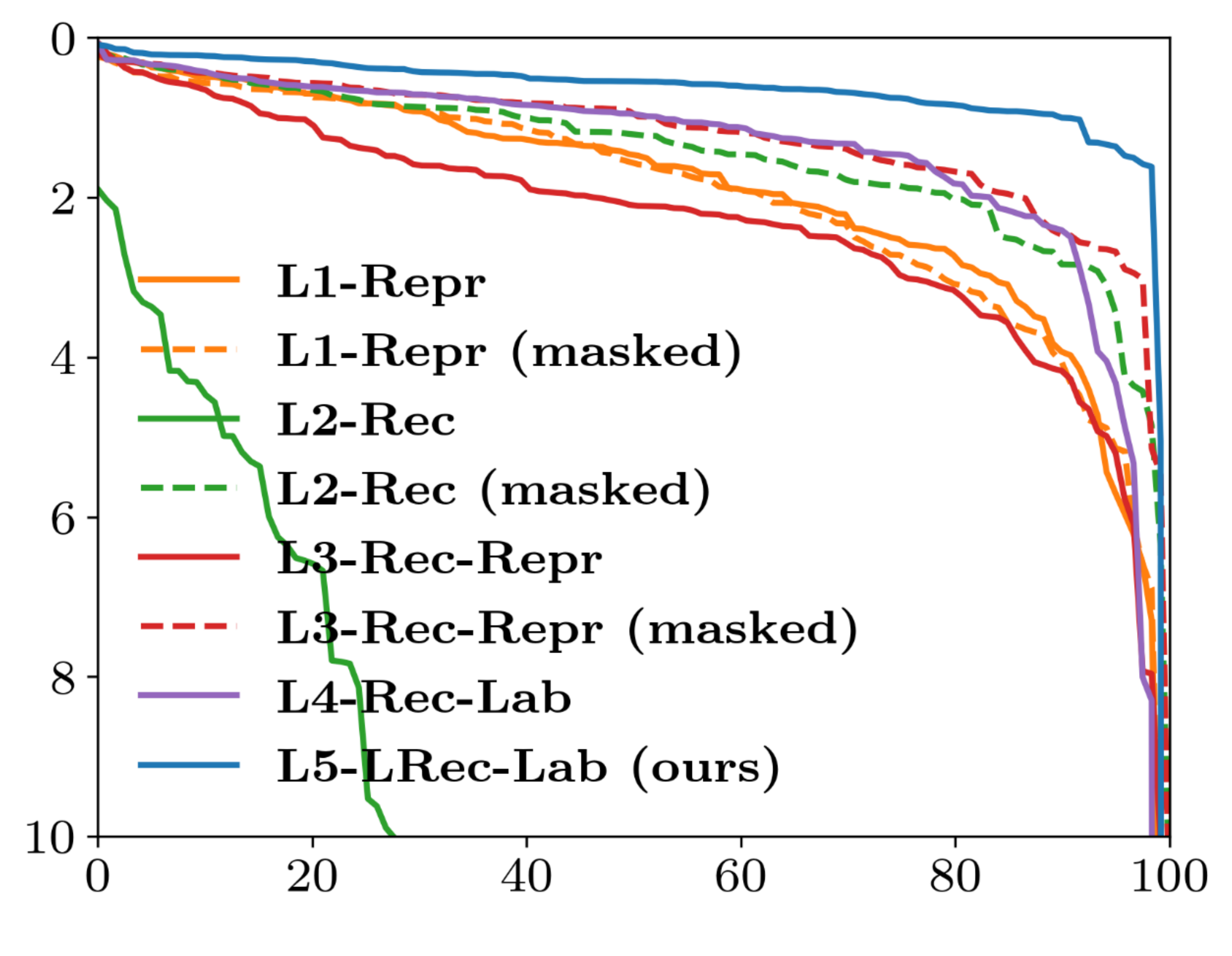}
\end{tabular}
   \caption{\small{Graph (a) plots the percentage of points (vertical axis) for which predicted scene coordinates reside within a given distance (horizontal axis) from a corresponding ground truth location. Our method (\textit{$\text{L5-LRec-Lab}$}) outperforms alternative approaches at predicting significantly more points with small euclidean distance. For distances larger than $4 m$, our method is less accurate due to error introduced by mis-predictions of globally unique instance labels. Graph (b) plots the evolution of the percentage of points (vertical axis) which reside within $0.5 m$ from the ground truth location as the number of training epochs (horizontal) increases. Methods which attempt at directly predicting 3D coordinates ($L2,L3,L4$) converge significantly slower. While they may eventually reach similar accuracy if the amount of epochs was increased significantly, such a solution would be impractical. Graphs (c) and (d) plot euclidean distance in meters (vertical axis) between predicted camera location and its ground truth location and angular error in degrees respectively. The predictions are sorted (horizontal axis) from the smallest on the left side to the largest on the right side.}} \vspace{-0.3cm}
\label{figpart0304roc}
\end{figure*}
\begin{figure*}[t]
   \includegraphics[width=1.0\linewidth]{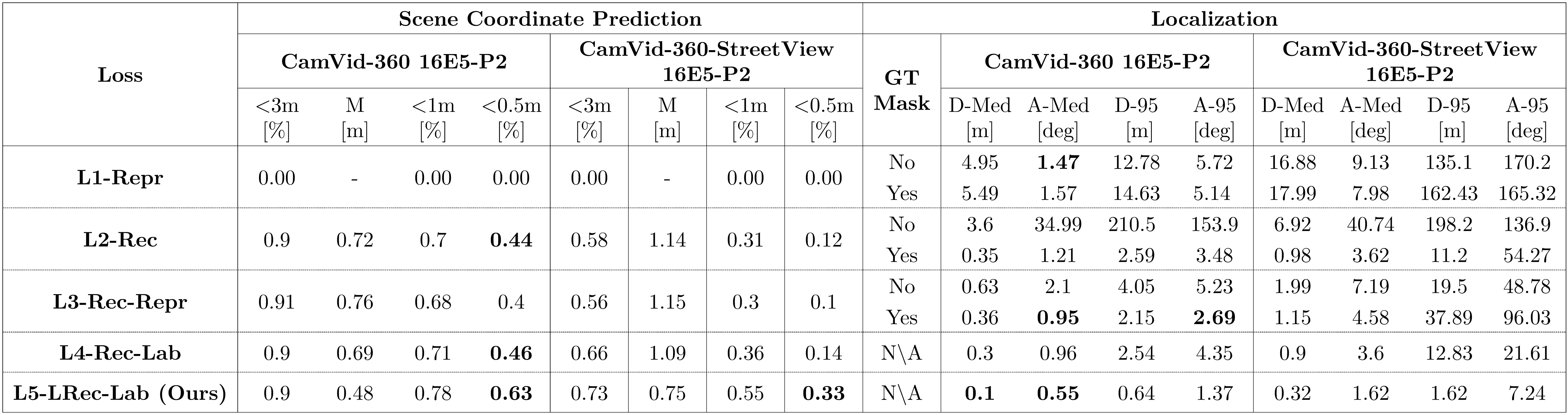} \vspace{-0.6cm}
       \caption{\small{This figure provides a quantitative evaluation of scene coordinate prediction and localisation performance for five different losses on CamVid-360 sequence 16E5-P2 and its StreetView counterpart. For the task of scene coordinate prediction percentages of pixels within $3 m$, $1 m$, and $0.5 m$ are reported together with average distance for pixels which are within 3m (column M). For the task of localisation median as well as 95th percentile angular error (A) and camera location distance from ground truth value (D) are reported. For methods which do not explicitly predict instance labels (\textit{$\text{L1-L3}$}), a result which is obtained by masking out pixels which do not belong to building instances (see column GT Mask) is reported for a fair evaluation.}} \vspace{-0.2cm}
\label{figcamvidvariationsquantitative}
\end{figure*}

\begin{figure*}[t]
    \includegraphics[angle=90,width=1.0\linewidth]{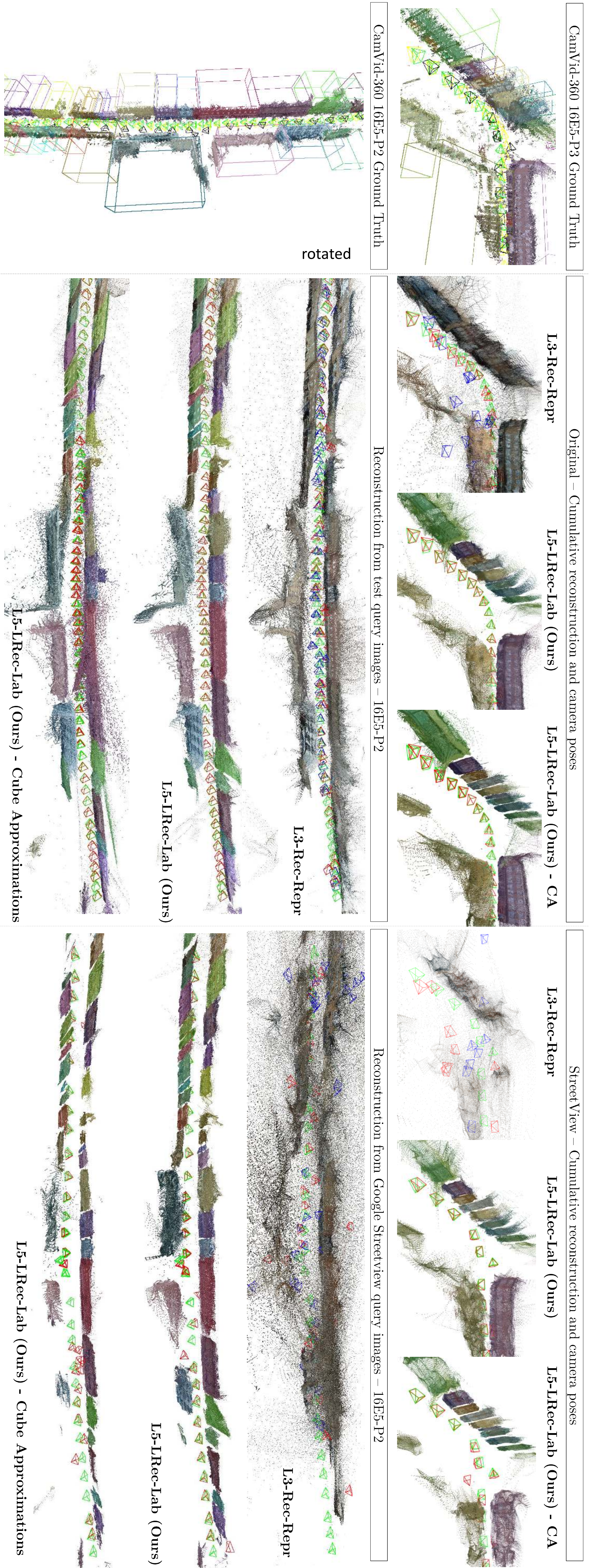} \vspace{-0.7cm}
   \caption{\small{The top left image of this figure displays the ground truth semantic point cloud as well as database (yellow) and query trajectories (green) for original CamVid-360 16E5-P3 sequence and our collected sequence from Google StreetView images (black). Two groups of three columns at the top show cumulative predicted 3D point-clouds (random sample of 1\% of total points) with accompanying ground truth camera poses (green) and predicted camera poses (red) for three different methods. Camera poses predicted by PoseNet~\cite{KendallC17} are marked in blue. Similar results are provided for sequence 16E5-P2 on the bottom part of the figure. Zoom in for a better view. Also see supplementary material.}} \vspace{-0.3cm}
\label{figcumulativequalitative}
\end{figure*}

\noindent \textbf{Training details.} Our network consists of a ResNet-50~\cite{HeZRS15ResNet} backbone followed by bilinear upsampling with skip layers analogously to FCN \cite{long2015fully}. ResNet-50~ implementation and initialization weights provided by the PyTorch~\cite{paszke2017automatic} repository are used. Strided convolution layers are replaced with dilated convolution in order to reduce the down-sampling inside the network. The models are trained for 3000 epochs unless stated otherwise, using a batch size of $14$ (for images of resolution $512\times256$) and the adam optimizer \cite{kingma2014adam}. The initial learning rate is set to 2e-4 and polynomial decrease \cite{liu2015parsenet,chen2018deeplab} is applied by multiplying the initial learning rate with $((1-\frac{step}{max\_steps})^{0.9})^2$ at each update step. An $L_2$ weight decay with factor 5e-4 is applied to all kernel weights and 2D-Dropout \cite{tompson2015efficient} with rate $0.5$ is used on top of the final convolutional layer. We train all networks on four GeForce GTX 1080 Ti GPUs.

\noindent \textbf{Evaluation protocol.} In this work, we provide quantitative and qualitative evaluation of the accuracy of both predicted scene coordinates as well as estimated 6-DoF camera poses. Scene coordinate regression is evaluated by measuring the percentange of points falling within $0.5 m$, $1.0 m$ and $3.0 m$ of corresponding ground truth targets. We also provide the average distance from ground truth coordinates for all points residing within $3 m$ of their targets. Points further away than 3m are excluded as they correspond to outliers which can obscure the true accuracy of the algorithms evaluated. The accuracy of 6-DoF camera pose estimation is evaluated by measuring median and 95-th percentile distance and angular errors between predicted and target cameras. See Figures~\ref{figcamvidvariationsquantitative} and~\ref{figcamvid360mainjoint} for example of results.

\vspace{-0.3cm}
\section{Experiments}\label{secexperiments} \vspace{-0.1cm}
Three types of experiments are performed in order to evaluate our proposed framework for joint re-localisation and scene understanding. In the first two sets of experiments we evaluate the quality of the scene coordinate prediction and localisation respectively. In the final set of experiments we explore the feasibility of performing localisation by using highly compact and fast-to-query maps which are made of cuboids approximating buildings.

\begin{figure*}[t]
   \includegraphics[width=1.0\linewidth]{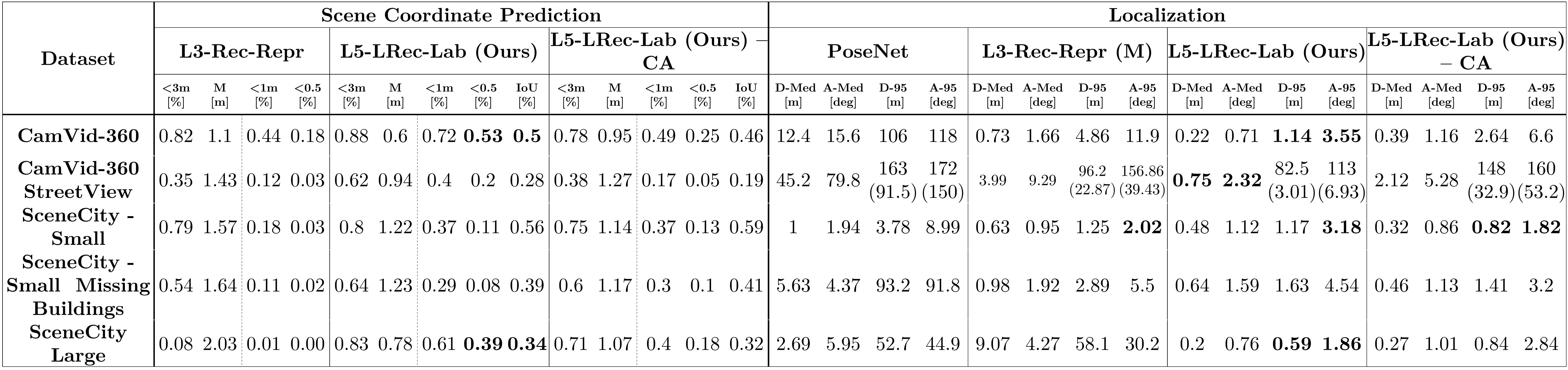} \vspace{-0.6cm}
   \caption{\small{This figure provides a quantitative evaluation of our approach on large datasets of CamVid-360 and SceneCity. Similar metrics are used as in Figure~\ref{figcamvidvariationsquantitative}. A method based on joint reconstruction and reprojection losses \textit{$\text{L3-Rec-Repr}$} is not able to accurately fit large maps. Our method demonstrates superior performance with more than $39\%$ of points predicted within $0.5 m$ on SceneCity Large. Relatively poorer reconstruction quality in SceneCity Small dataset (compared to SeneCity Large) can be explained by a higher density of tall buildings as the accuracy of the 3D points drops significantly for the tops of buildings. Our method \textit{$\text{L5-Rec-Repr (Ours)}$} outperforms both PoseNet~\cite{KendallC17} and \textit{$\text{L3-Rec-Repr}$} in all experiments with an exception of the angular distance error on Small SceneCity dataset due to the effect of reprojection loss component in $L3$. However a version of our method which uses approximate 3D maps outperforms $L3$. This can be explained by cuboids being a good approximation of artificial buildings. Also note that numbers in brackets correspond to 80th percentile distance and angular errors for CamVid-360 StreetView dataset.
   }} \vspace{-0.5cm}
\label{figcamvid360mainjoint}
\end{figure*}

\noindent \textbf{Scene coordinate regression.} Firstly, we compare five CNNs trained with different losses and evaluate their performance on the task of scene coordinate regression on a subsequence 16E5-P2 of CamVid-360~\cite{BudvytisSC18} dataset. The first loss \textit{$\text{L1-Repr}$} minimizes a reprojection error of points on a spherical image plane: $\textit{\text{L1-Repr}}(S,S^{gt})=\sum^{}_{p \in P }||\frac{R^{T}S_{p}+T}{||R^{T}S_{p}+T||}-V_{p}||$. Here $R$ and $T$ are ground truth camera rotation and translation matrices, $P$ - a set of all pixels in an image, $S_{p}$ - predicted scene coordinates for a pixel $p$ and $V_{p}$ is a vector pointing to pixel $p$ projection on a spherical image. It is a straight-forward adaptation a reprojection loss used in~\cite{BrachmannR18} from planar images to spherical images. It is also equivalent to a loss proposed in~\cite{liangularscr18}. The second loss $\textit{\text{L2-Rec}}=\sum^{}_{p \in M^{gt}}||S_{p}-S^{gt}_{p}||$ directly minimises the euclidean distance between predicted scene coordinates $S_{p}$ and ground truth scene coordinates $S^{gt}_{p}$ for all pixels for which ground truth coordinates are available - set $M^{gt}$. Depending on choices of learning rates and relative loss weighting parameters the first two stages of the approach of~\cite{BrachmannR18} can be viewed as a mixture of both aforementioned losses. We approximate this work by loss $\textit{\text{L3-Rec-Repr}}=\alpha\textit{\text{L1-Rep}}+(1-\alpha)\textit{\text{L2-Rec}}$, where $\alpha$ is set empirically to $0.02$. Note that we do not use the third stage of differentiable pose prediction and use a classical method~\cite{lepetitepnp} instead. Also note that authors of~\cite{BrachmannR18} report only a small advantage of using this stage at a cost of the lack of convergence on Street scene of Cambridge Landmarks~\cite{PoseNetKendallGC15} dataset. The final two losses considered are \textit{$\text{L4-Rec-Lab}$} and \textit{$\text{L5-LRec-Lab}$}. The former combines a standard cross entropy loss used for semantic segmentation and a reconstruction loss \textit{$\text{L2}$}, whereas the latter combines a cross entropy loss with reconstruction loss in local whitened coordinate space. We empirically set relative weighting between cross entropy loss and reconstruction losses to $0.1$ and $0.5$ respectively. As shown in Figures~\ref{fig3dqualitative}(a),~\ref{figpart0304roc}(a) and~\ref{figcamvidvariationsquantitative} reprojection loss (\textit{$\text{L1-Repr}$}) alone does not enable a CNN to recover accurate geometry and instead predict a point cloud of an approximately spherical shape. In contrast, using methods which directly predict scene coordinates (\textit{$\text{L2}$}, \textit{$\text{L3}$}, \textit{$\text{L4}$}) lead to high accuracies with more than $40\%$ of points residing within $0.5 m$ of their ground truth location. Our method \textit{$\text{L5-LRec-Lab}$} outperforms the alternatives by more than $17\%$. It is due to a faster convergence at train time which is caused by a simpler optimization task resulting from the separation of object center and local coordinate prediction (see Figure~\ref{figpart0304roc}(b)). The difference in performance between the aforementioned methods becomes even bigger when larger maps such as full CamVid-360 or artificial cities are considered as shown in Figures~\ref{fig3dqualitative}(b),~\ref{figcumulativequalitative} and~\ref{figcamvid360mainjoint}.

\begin{figure*}[t]
   \includegraphics[width=1.0\linewidth]{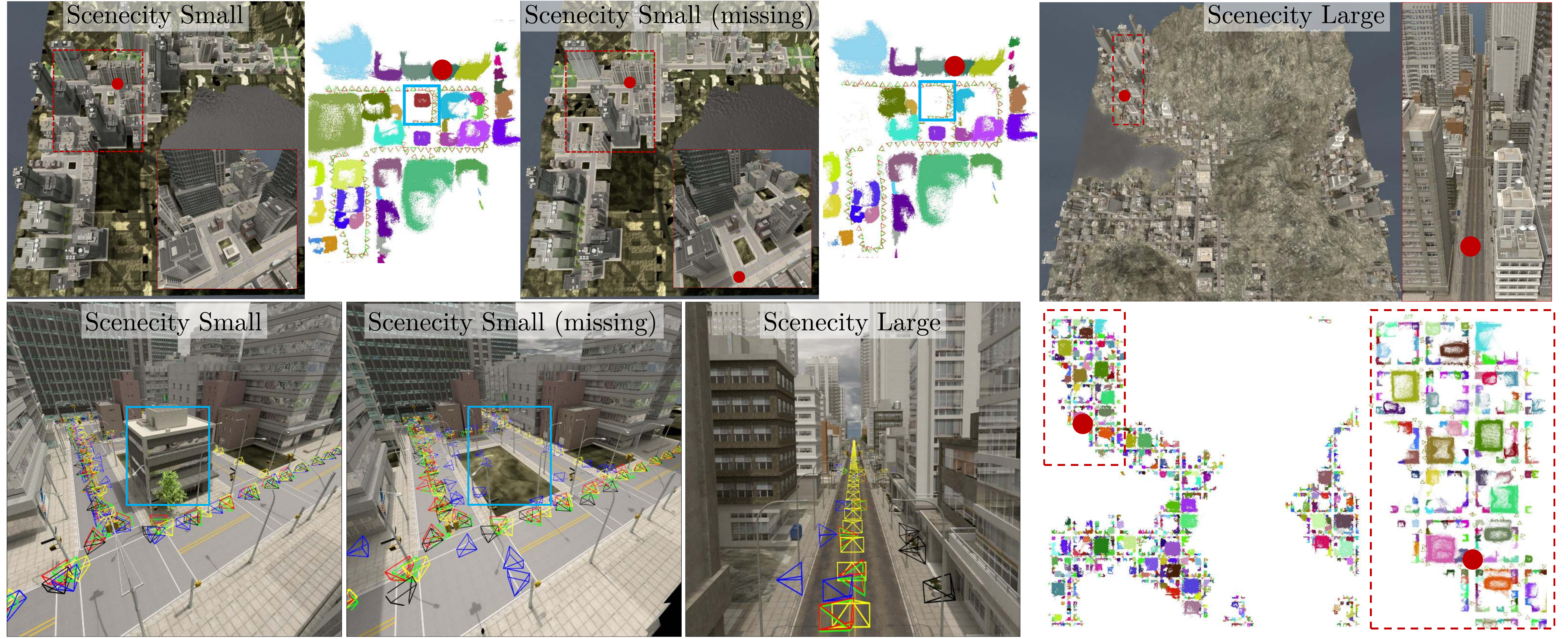} \vspace{-0.5cm}
   \caption{\small{The top row of this figure shows artificial city maps using a top-view orthographic projection. Each city view has a region zoomed in and visualised from a different angle and a corresponding 3D point cloud obtained by accumulating 3D points predicted from test images. Examples of missing buildings are marked in blue rectangles. Three images at the bottom left illustrate the view seen by a camera marked in red dot. They show camera poses of ground truth database (yellow), ground truth query (green), \textit{$\text{L3-Rec-Repr}$} (black), PoseNet~\cite{KendallC17} (blue) and \textit{$\text{L5-LRec-Lab (Ours)}$} (red). Zoom in for a better view. Also see supplementary material.
   }} \vspace{-0.5cm}
\label{figreconstructedpaths}
\end{figure*}

\noindent \textbf{Localisation.} As with scene coordinate regression we first evaluate localisation accuracy of various methods on the sequence $\text{16E5-P2}$ of CamVid-360~\cite{BudvytisSC18} dataset in detail. We then follow by experiments on full length sequences of CamVid-360 and two artificial cities. It can be seen in  Figures~\ref{figpart0304roc}(c) and (d) that CNN trained using \textit{$\text{L1-Repr}$} loss shows a poor performance in estimating 3D location, but a relatively low angular error. \textit{$\text{L2-Rec}$} and \textit{$\text{L3-Rec-Repr}$} perform poorly at both tasks if pixels not belonging to building instances are not masked out at test time. Note that we use ground truth masks in order to evaluate the upper bound of the performance of both methods. Directly predicting semantic scene coordinates (loss \textit{$\text{L4-Rec-Lab}$}) produces performance similar to masked versions of \textit{$\text{L2}$} and \textit{$\text{L3}$} as its localisation accuracy is limited by the accuracy of 3D coordinates predicted. Hence it is not surprising that our proposed method  based on predicting local object coordinates (\textit{$\text{L5-LRec-Lab}$}) outperforms all  the alternative methods at both angular error and camera location distance error significantly. Similar trends are observed on large experiments as reported quantitatively in Figure~\ref{figcamvid360mainjoint} and qualitatively in Figures~\ref{figcumulativequalitative} and~\ref{figreconstructedpaths}. Also note that while PoseNet~\cite{KendallC17} (a standard setup\footnote{Note that in order to use PoseNet~\cite{KendallC17} on equirectangular images an explicit rotation augmentation of the camera pose needs to be performed for each crop. We limited crops to a horizontal shift only which corresponds to the rotation of the camera around its axis.} with geometric reprojection error and ResNet-50 encoder), adapted to performing on panoramic images, shows a seemingly competitive performance on Small and Large SceneCity data, its performance drops on CamVid-360 dataset. It can be explained by PoseNet~\cite{KendallC17} sensitivity to overfitting to the training images as they are obtained from a single video as opposed to a diverse set of images. This is also supported by a significant drop in accuracy on Small SceneCity images with missing buildings. 

%\newpage
\noindent \textbf{Localisation in approximate maps.} In the final set of experiments we explore the alternative of predicting a 6-DoF pose from simplified approximate 3D maps. As expected, on CamVid-360 dataset, scene coordinate and camera pose prediction is of lower accuracy than the one of the network trained using precise 3D map as shown in Figures~\ref{figcumulativequalitative} and~\ref{figcamvid360mainjoint}. However on the Small SceneCity data, this network outperforms \textit{$\text{L5-LRec-Lab (Ours)}$}. This can be explained by cuboids being a good approximation for buildings in artificial cities. Moreover higher performance in localisation is obtained than competing approaches of PoseNet~\cite{KendallC17} and \textit{$\text{L3-3D-Repr}$} in all experiments. This is a highly encouraging result which in the future may alleviate the need of a computationally expensive step of building 3D point clouds of cities.
%------------------------------------------------------------------------

\vspace{-0.3cm}
\section{Conclusions} \vspace{-0.1cm}

In this work we presented a novel approach for a large scale joint semantic re-localisation and scene understanding via globally unique instance coordinate regression. To the best of our knowledge this is the first work to demonstrate that scene coordinate regression framework can be used for a large scale localisation. We achieve this by separating the task of scene coordinate prediction into object instance segmentation and local coordinate prediction. This significantly speeds up the convergence of scene coordinate regression networks and allows to achieve high scene coordinate prediction and localisation accuracies. 

\bibliography{egbibv9}
\end{document}